\documentclass[11pt,a4paper]{article}
\usepackage[hyperref]{acl2021}
\usepackage{times}
\usepackage{latexsym}
\usepackage{array}
\usepackage{adjustbox}
\usepackage{amsmath, amsfonts}

\usepackage{microtype}
\usepackage{stackengine}

\aclfinalcopy % Uncomment this line for the final submission
\def\aclpaperid{671} %  Enter the acl Paper ID here

\newcommand\R{\mathbb{R}}

\title{Convolutions and Self-Attention: Re-interpreting Relative Positions in Pre-trained Language Models}

\author{Tyler A. Chang$^{1,3}$, \quad Yifan Xu$^1$, \quad Weijian Xu$^2$, \quad Zhuowen Tu$^{1,2,3}$ \\
$^1$Department of Cognitive Science \\
$^2$Department of Computer Science and Engineering \\
$^3${Halıcıoğlu} Data Science Institute \\
University of California San Diego \\
{\texttt{$\{$tachang, yix081, wex041, ztu$\}$@ucsd.edu}}
}

\date{}

\begin{document}
\maketitle
\begin{abstract}
In this paper, we detail the relationship between convolutions and self-attention in natural language tasks. We show that relative position embeddings in self-attention layers are equivalent to recently-proposed dynamic lightweight convolutions, and we consider multiple new ways of integrating convolutions into Transformer self-attention.
Specifically, we propose composite attention, which unites previous relative position embedding methods under a convolutional framework.
We conduct experiments by training BERT with composite attention, finding that convolutions consistently improve performance on multiple downstream tasks, replacing absolute position embeddings. To inform future work, we present results comparing lightweight convolutions, dynamic convolutions, and depthwise-separable convolutions in language model pre-training, considering multiple injection points for convolutions in self-attention layers.
\end{abstract}

\section{Introduction}
In recent years, Transformer-based language models have brought dramatic improvements on a wide range of natural language tasks \citep{brown-etal-2020-language,devlin-etal-2019-bert}.
The central innovation of Transformer architectures is the self-attention mechanism \citep{vaswani-etal-2017-attention}, which has grown beyond NLP, extending into domains ranging from computer vision \citep{dosovitskiy2021image} and speech recognition \citep{dong2018speech} to reinforcement learning \citep{parisotto-etal-2020-stabilizing,touvron-etal-2020-training}.

In computer vision, self-attention and convolutions have been combined to achieve competitive results for image classification \citep{bello2019attention}. 
Similarly, researchers in NLP have begun integrating convolutions into self-attention for natural language tasks.
Recent work has shown initial success adding convolutional modules to self-attention in pre-trained language models \citep{jiang-etal-2020-convbert}, or even replacing self-attention entirely with dynamic convolutions \citep{wu-etal-2019-pay}.
These successes defy theoretical proofs showing that multi-headed self-attention with relative position embeddings is strictly more expressive than convolution \citep{cordonnier-etal-2020-relationship}.
To identify why convolutions have been successful in NLP, we seek to isolate the differences between self-attention and convolution in the context of natural language.

In this work, we formalize the relationship between self-attention and convolution in Transformer encoders by generalizing relative position embeddings, and we identify the benefits of each approach for language model pre-training. We show that self-attention is a type of dynamic lightweight convolution, a data-dependent convolution that ties weights across input channels \citep{wu-etal-2019-pay}. Notably, previous methods of encoding relative positions \citep{shaw-etal-2018-self,raffel-etal-2020-exploring} are direct implementations of lightweight convolutions.
Under our framework, the benefits of convolution come from an ability to capture local position information in sentences.

Then, we propose composite attention, which applies a lightweight convolution that combines previous relative position embedding methods. We find that composite attention sufficiently captures the information provided by many other convolutions. To validate our framework, we train BERT models that integrate self-attention with multiple convolution types, evaluating our models on the GLUE benchmark \citep{wang-etal-2018-glue}.
All of our convolutional variants outperform the default model, demonstrating the effectiveness of convolutions in enhancing self-attention for natural language tasks.
Our empirical results provide evidence for future research integrating convolutions and self-attention for NLP.

\begin{figure*}
    \centering
    % left, bottom, right, top
    \adjustbox{trim=1cm 2cm 0cm 0cm}{%
    \includegraphics[width=18cm]{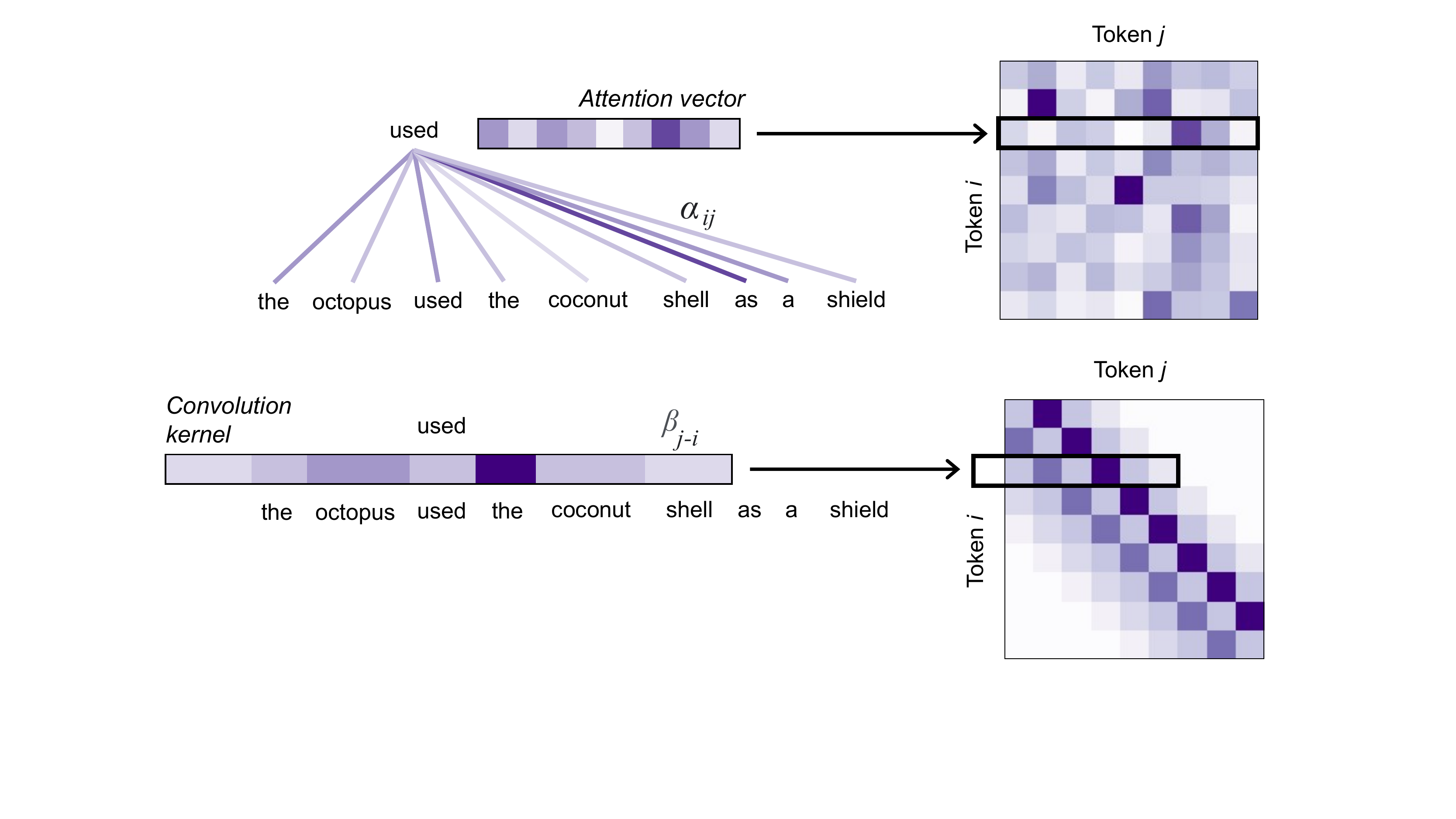}%
    }
    \caption{Generating attention maps using standard self-attention (top) and fixed lightweight convolution (bottom). Attention weights $\alpha_{ij}$ are analogous to convolution kernel weights $\beta_{j-i}$.}
    \label{conv-attention-fig}
\end{figure*}

\section{Self-attention and lightweight convolutions}
First, we outline the relationship between self-attention and convolutions.
Specifically, we show that a self-attention operation can be viewed as a dynamic lightweight convolution, a depthwise convolution that ties weights along channels \citep{wu-etal-2019-pay}. We then isolate the differences between self-attention and lightweight convolutions, highlighting the benefits of each approach in language models.

\subsection{Self-attention}
In a Transformer self-attention layer, inputs $\textbf{x}_1, ..., \textbf{x}_n \in \R^d$ are projected to corresponding queries, keys, and values by linear transformations $W^Q, W^K, W^V \in \R^{d \times d_h}$ for each attention head, projecting into the head dimension size $d_h$.
Output vectors $\textbf{y}_1, ..., \textbf{y}_n \in \R^d$ are linear combinations of values, concatenating all attention heads.
Value weights (before softmaxing) are determined by:
\begin{equation}
\alpha_{ij} = \frac{(\textbf{x}_i W^Q) (\textbf{x}_j W^K)^T}{\sqrt{d_h}}.
\end{equation}
Intuitively, $\alpha_{ij}$ represents the attention that token $i$ pays to token $j$, incorporating the value $\textbf{x}_j W^V$ into the resulting vector $\textbf{y}_i$.
From the attention scores between various tokens $i$ and $j$, an attention map of $\alpha_{ij}$ is produced (see Figure \ref{conv-attention-fig}).

\subsection{Lightweight convolutions}
In contrast, a standard one-dimensional convolution slides a kernel of weights along the input sequence; each feature in each output representation $\textbf{y}_i$ is a weighted sum of all features (called “channels”) in the surrounding $\textbf{x}_i$.
To save parameters, it is common to consider depthwise convolutions where each channel $c$ in $\textbf{y}_i$ is a weighted sum only of the features in channel $c$ for the surrounding $\textbf{x}_i$. Formally, each entry of $\textbf{y}_i$ can be written as:
\begin{equation}
y_{i,c} = \sum_{-k \leq j-i \leq k} \beta_{j-i,c} \hspace{0.1cm} x_{j, c}
\end{equation}
where $k$ is the kernel size in each direction.
Each scalar $\beta_{j-i, c}$ represents the attention paid to relative position $j-i$ for channel $c$.
To further simplify depthwise convolutions for use in language models, \citet{wu-etal-2019-pay} propose lightweight convolutions, which tie weights $\beta_{j-i,c}$ along all channels $c$.
As a result, the lightweight convolution contains only $2k+1$ weights, one scalar $\beta_{j-i}$ for each relative position considered.
Then, each $\textbf{y}_i$ is a linear combination of surrounding $\textbf{x}_i$:
\begin{equation}
\textbf{y}_i = \sum_{-k \leq j-i \leq k} \beta_{j-i} \hspace{0.1cm} \textbf{x}_j
\end{equation}
Importantly, we can then consider each $\beta_{j-i}$ as an attention weight analogous to self-attention, representing the attention that token $i$ pays to token $j$.
The lightweight convolution produces an attention map of $\beta_{j-i}$ as visualized in Figure \ref{conv-attention-fig}.

Finally, furthering the similarity between lightweight convolutions and self-attention, \citet{wu-etal-2019-pay} propose dynamic lightweight convolutions, which dynamically compute relative weights $\beta_{j-i}$ based on individual input tokens.
In other words, each row in Figure \ref{conv-attention-fig} has relative weights determined dynamically based on the input token $\textbf{x}_i$ for that row.
Because attentions for relative positions are no longer fixed across rows, the attention map in Figure \ref{conv-attention-fig} achieves similar flexibility to standard self-attention.

\subsection{Self-attention vs. convolution}
We have shown that both self-attention and lightweight convolution compute linear combinations of token representations, but we now isolate the differences between the two approaches.
Perhaps most importantly, the two methods assign attention scores $\alpha_{ij}$ and $\beta_{j-i}$ in fundamentally different ways.

Self-attention computes $\alpha_{ij}$ based on the dot product between query $i$ and key $j$, ignoring the relative position between $i$ and $j$.
In this way, self-attention layers model interactions exclusively between token representations.
If the tokens are arbitrarily shuffled in a standard self-attention layer, the output for each token is unchanged.
All position information is injected before the first self-attention layer in the form of absolute position embeddings.

In contrast, dynamic lightweight convolutions assign attention scores directly to relative positions.
This allows convolutions to directly integrate relative position information without relying on absolute positions.
Thus, convolutions could be better at capturing local information in sentences.
However, convolutions alone are limited in their ability to model interactions between tokens because they lack the query-key mechanism central to standard self-attention.
In future sections, we consider methods of integrating the two approaches.

\section{Integrating lightweight convolutions}
Previous work has sought to integrate local information into global self-attention.
This can be achieved by restricting the range of self-attention to nearby tokens, or by incorporating relative position information into attention maps \citep{hofstatter-etal-2020-local,raganato-etal-2020-fixed,wei-etal-2021-position}.
Notably, \citet{shaw-etal-2018-self} introduced relative position embeddings, which inspired similar embeddings in models such as Transformer-XL and XLNet \citep{dai-etal-2019-transformer,yang-etal-2019-xlnet}.
In this section, we show that several previous methods of encoding relative positions are direct implementations of lightweight convolutions.

\subsection{Relative embeddings as lightweight convolutions}
First, the simplest way to combine self-attention with lightweight convolution is to generate a standard attention map, then add the attention map generated by a lightweight convolution. 
Given a fixed lightweight convolution, this results in attention scores as follows:
\begin{equation}\label{eq:convfixed}
\alpha_{ij} = \frac{(\textbf{x}_i W^Q) (\textbf{x}_j W^K)^T}{\sqrt{d_h}} + \beta_{j-i}
\end{equation}
This is exactly the relative position term used in T5 \citep{raffel-etal-2020-exploring} and TUPE \citep{ke-etal-2021-rethinking}.

We further consider a dynamic lightweight convolution, where the $\beta_{j-i}$ weights are computed by passing the query through a linear feedforward layer $W^C \in \R^{d_h \times (2k+1)}$ \citep{wu-etal-2019-pay}.\footnote{\citet{wu-etal-2019-pay} generate dynamic lightweight convolutions based on the entire query layer (dimension size $d$). In our work, we generate convolutions based on queries for individual attention heads (dimension size $d_h$), to be consistent with the relative embeddings in \citet{shaw-etal-2018-self}.}
Because $W^C$ is linear, each weight $\beta_{j-i}$ is equal to the dot product between the query and the $(j-i)$ column of $W^C$. 
We then obtain attention scores:
\[
\alpha_{ij} = \frac{(\textbf{x}_i W^Q) (\textbf{x}_j W^K)^T}{\sqrt{d_h}} + (\textbf{x}_i W^Q)(W^C_{j-i})^T
\]
If we scale the dynamic lightweight convolution term according to the head dimension size, we obtain precisely the relative embeddings proposed in \citet{shaw-etal-2018-self}:
\begin{equation}\label{eq:convq}
\alpha_{ij} = \frac{(\textbf{x}_i W^Q) (\textbf{x}_j W^K + W^C_{j-i})^T}{\sqrt{d_h}}
\end{equation}
Under this interpretation, Shaw's relative embeddings are essentially identical to the dynamic lightweight convolutions used in \citet{wu-etal-2019-pay}.
In both formulations, relative position weights are computed as dot products between the query and a learned relative position embedding.
Previous work has considered relative positions in language models independently from convolutions, but our derivations suggest that the underlying mechanisms may be the same.

\setlength\tabcolsep{3pt}
\begin{table*}[t]
    \centering
    \small
    \renewcommand{\arraystretch}{1.2}
    \begin{tabular}{|>{\raggedright}p{4cm}|>{\raggedleft}p{1.12cm}|>{\raggedleft}p{0.8cm}>{\raggedleft}p{0.9cm}>{\raggedleft}p{0.9cm}>{\raggedleft}p{0.9cm}>{\raggedleft}p{0.8cm}>{\raggedleft}p{0.8cm}>{\raggedleft}p{0.75cm}>{\raggedleft}p{0.7cm}>{\raggedleft}p{0.7cm}|r|}
        \hline
        \textbf{Lightweight convolution \newline type, BERT-small} & \textbf{Params} & CoLA & MNLI-m & MNLI-mm & MRPC & QNLI & QQP & RTE & SST & STS & \textbf{GLUE} \\
        \hline
         No convolution & 13.41M & 13.9 & 73.2 & 71.8 & 77.9 & 80.7 & 74.5 & 62.0 & 81.9 & 79.3 & 68.4 \\
        No convolution  + abs position$^*$ & 13.43M & 30.8 & 76.1 & 75.9 & 80.4 & 78.5 & 74.4 & 62.2 & 85.1 & 76.8 & 71.1 \\
        Fixed (\citealt{raffel-etal-2020-exploring}) & 13.42M & \textbf{42.1} & 77.2 & 76.3 & 83.8 & 82.7 & 75.9 & 64.4 & 87.1 & 81.4 & 74.5 \\
        Dynamic (\citealt{shaw-etal-2018-self}) & 13.43M & 39.1 & \textbf{78.4} & \textbf{77.4} & 83.8 & \textbf{83.4} & 77.5 & 64.4 & 87.3 & 81.4 & 74.7 \\
        \hline
        Composite (Equation \ref{eq:convq-convfixed}; ours) & 13.43M & 40.4 & 78.2 & \textbf{77.4} & \textbf{85.0} & 83.3 & \textbf{77.7} & \textbf{64.7} & \textbf{87.8} & \textbf{82.1} & \textbf{75.2} \\
        \hline
        \multicolumn{12}{c}{ } \\
        \hline
        \textbf{Lightweight convolution \newline type, BERT-base} & \textbf{Params} & CoLA & MNLI-m & MNLI-mm & MRPC & QNLI & QQP & RTE & SST & STS & \textbf{GLUE} \\
        \hline
        No convolution + abs position$^*$ & 108.82M & 50.3 & \textbf{82.0} & \textbf{81.2} & 85.0 & 84.6 & 78.6 & 68.9 & 91.4 & 84.9 & 78.5 \\
        Fixed (\citealt{raffel-etal-2020-exploring}) & 108.73M & 50.0 & 81.5 & 80.5 & \textbf{85.6} & \textbf{86.0} & 78.5 & 68.9 & 91.4 & 84.9 & 78.6 \\
        Dynamic (\citealt{shaw-etal-2018-self}) & 108.74M & \textbf{50.9} & 81.6 & 80.5 & 84.6 & 85.3 & 78.5 & 69.5 & \textbf{91.6} & 84.8 & 78.6 \\
        \hline
        Composite (Equation \ref{eq:convq-convfixed}; ours) & 108.74M & 50.4 & 81.6 & 80.8 & 85.4 & 85.1 & \textbf{78.7} & \textbf{69.7} & 91.2 & \textbf{85.7} & \textbf{78.7} \\
        \hline
    \end{tabular}
    \caption{GLUE test set performance for models with lightweight convolutions added to self-attention. Columns indicate scores on individual GLUE tasks; the final GLUE score is the average of individual task scores. $^*$ denotes the default BERT model.}
    \label{tab:lightweight}
\end{table*}

\subsection{Composite attention and lightweight convolution experiments}\label{section-lightweight-experiments}
To validate lightweight convolutions in combination with self-attention, we pre-trained and evaluated BERT-small models \citep{devlin-etal-2019-bert,clark-etal-2020-electra} that incorporated lightweight convolutions.

\paragraph{Pre-training}
To maximize similarity with \citet{devlin-etal-2019-bert}, we pre-trained models on the BookCorpus \citep{zhu-etal-2015-aligning} and WikiText-103 datasets \citep{merity-etal-2017-pointer} using masked language modeling.
Small models were pre-trained for 125,000 steps, with batch size 128 and learning rate 0.0003.
Full pre-training and fine-tuning details are outlined in Appendix \ref{app:hyperparams}.\footnote{Code is available at \url{https://github.com/mlpc-ucsd/BERT_Convolutions}, built upon the Huggingface Transformers library \citep{wolf-etal-2020-transformers}.}

\paragraph{Evaluation}
Models were evaluated on the GLUE benchmark, a suite of sentence classification tasks including natural language inference (NLI), grammaticality judgments, sentiment classification, and textual similarity \citep{wang-etal-2018-glue}.
For each task, we ran ten fine-tuning runs and used the model with the best score on the development set.
We report scores on the GLUE test set.
Development scores and statistics for all experiments are reported in Appendix \ref{app:dev-results}.

\paragraph{Models}
We trained two baseline models, a default BERT-small with standard absolute position embeddings, and a BERT-small with no position information whatsoever.
Then, we trained models with fixed lightweight convolutions (Equation \ref{eq:convfixed}; \citealt{raffel-etal-2020-exploring}), and dynamic lightweight convolutions that generated convolution weights based on each query (i.e. using relative embeddings, Equation \ref{eq:convq}; \citealt{shaw-etal-2018-self}).

Finally, we propose composite attention, which simply adds dynamic lightweight convolutions to fixed lightweight convolutions, resulting in attention scores $\alpha_{ij}$ as follows:

\begin{equation}\label{eq:convq-convfixed}
\begin{aligned} % Just using this to add a space below.
\underbrace{\textstyle \frac{(\textbf{x}_i W^Q) (\textbf{x}_j W^K)^{T}}{\sqrt{d_h}}}_{\textrm{Self-attention}} + \underbrace{\textstyle \frac{(\textbf{x}_i W^Q)(W^C_{j-i})^T}{\sqrt{d_h}}}_{\substack{\textrm{Dynamic convolution} \\ \textrm{(relative embeddings)}}} + \underbrace{{\beta_{j-i}}}_{\substack{\textrm{Fixed} \\ \textrm{convolution}}} \\ \\
\end{aligned} % \addstackgap[6pt]
\end{equation}
Intuitively, composite attention has the flexibility of dynamic lightweight convolutions, while still allowing models to incorporate relative positions directly through fixed lightweight convolutions.
Alternatively, composite attention can be interpreted as adding a fixed bias term to relative position embeddings.

All of our experiments used a convolution kernel size of 17, or eight positions in each direction, a mid-range value that has been found to work well for both relative positions and convolution in language models \citep{huang-etal-2020-improve,jiang-etal-2020-convbert,shaw-etal-2018-self}.
As in \citet{shaw-etal-2018-self}, relative embeddings $W^C_{j-i}$ shared weights across heads.
Unless stated otherwise, models used no absolute position embeddings.

For completeness, we also considered dynamic lightweight convolutions based on the key (as opposed to the query).
In contrast to query-based lightweight convolutions, key-based convolutions allow each token to dictate which relative positions should pay attention to it, rather than dictating which relative positions it should pay attention to.
Referring to the visualization in Figure \ref{conv-attention-fig}, key-based dynamic convolutions correspond to columns instead of rows.
These key-based dynamic lightweight convolutions are the same as the relative embeddings proposed in \citet{huang-etal-2020-improve}, but they are now formulated as dynamic lightweight convolutions.

\subsection{Lightweight convolution results}
GLUE test set results are presented in Table \ref{tab:lightweight}.

\paragraph{Lightweight convolutions consistently improved performance.}
Notably, even the fixed lightweight convolution was sufficient to replace absolute position embeddings, outperforming the default BERT-small model.
This indicates that even na\"{i}ve sampling from nearby tokens can be beneficial to language model performance.

\paragraph{Dynamic convolutions provided further improvements.}
When the lightweight convolutions were generated dynamically based on token queries, the models outperformed the default model by even larger margins.
This improvement over fixed lightweight convolutions suggests that different tokens find it useful to generate different lightweight convolutions, paying attention to different relative positions in a sentence.

\paragraph{Composite attention performed the best.}
Combining fixed lightweight convolutions with dynamic lightweight convolutions proved an effective strategy for encoding relative positions.
Although composite attention is simply a combination of \citet{shaw-etal-2018-self} and \citet{raffel-etal-2020-exploring}'s relative position embeddings, it validates convolution as a viable method of encoding relative positions in self-attention.

\paragraph{Key-based dynamic convolutions provided no additional benefit.}
When we generated an additional lightweight convolution based on keys, the model performed worse than composite attention alone (GLUE 74.0 compared to 75.2).
This result clarifies the findings of \citet{huang-etal-2020-improve}, who reported only small improvements from query and key-based relative position embeddings for a subset of the GLUE tasks.

\begin{figure}[!htp]
    \centering
    % left, bottom, right, top
    \includegraphics[width=5cm]{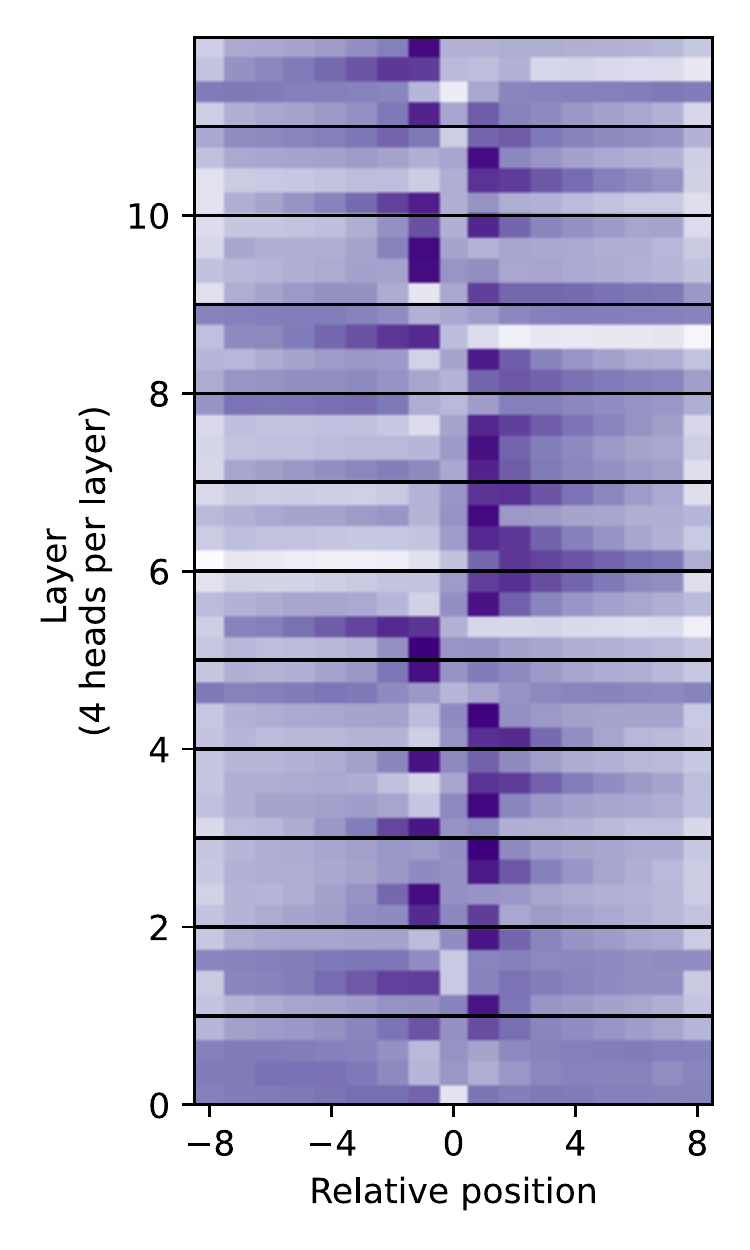}%
    \vspace{-2mm}
    \caption{Learned convolution kernel weights $\beta_{j-i}$ for the fixed lightweight convolution (Equation \ref{eq:convfixed}).}
    \label{fig:convfixed-weights}
    \vspace{-2mm}
\end{figure}

\paragraph{Grammaticality judgments were particularly sensitive to position information.}
On the CoLA task (the corpus of linguistic acceptability; \citealt{warstadt-etal-2019-neural}), there was a dramatic performance drop when absolute position embeddings were removed.
However, when any type of lightweight convolution was added, performance improved even over the baseline established by absolute positions.
The pronounced effects of local position information on the CoLA task support the intuitive hypothesis that local dependencies are particularly important for grammaticality judgments.
This result also suggests that convolutions could be beneficial to more local tasks (e.g. token-level tasks) along with sentence classification tasks. 

\subsection{Interpreting lightweight convolutions}
To better understand how lightweight convolutions improve language models, we visualized the learned lightweight convolution kernel weights in Figure \ref{fig:convfixed-weights}.
Qualitatively, the kernels exhibited specific types of patterns:
\vspace{-0.2em}
\begin{itemize}
    \setlength\itemsep{0.1em}
    \item Paying particular attention to the previous or next token.
    \item Paying graded attention either to past or future tokens, dictated by how far the target token is from the present token.
\end{itemize}
\vspace{-0.2em}
These observations support the assumption that nearby tokens are relevant to the interpretation of the current token.
They also align with the findings of \citet{voita-etal-2019-analyzing}, who identified ``positional'' attention heads that focus primarily on the next or previous token.
From this perspective, lightweight convolutions allow language models to explicitly represent nearby tokens' positions.

Interestingly, we also found that some kernels paid fairly uniform attention to all tokens, even decreasing attention to nearby and adjacent tokens.
It is likely that these attention heads focused on more global information, relying on the query-key attention mechanism rather than the convolution.

\subsection{BERT-base models}
To thoroughly assess the impact of composite attention on pre-trained language models, we trained full-sized BERT models for 1M steps each, replicating our BERT-small experiments.
Pre-training details are outlined in Appendix \ref{app:hyperparams}.

Results are presented in Table \ref{tab:lightweight}.
Differences between models decreased substantially for full sized models, and the relative performances of different approaches varied across tasks.
Our results suggest that relative position information is more useful for smaller or more data-limited models; extending the benefits of convolutions robustly from small models to larger models is an important direction for future research.
That said, even in the larger models, composite attention slightly outperformed the other position embedding methods in overall GLUE score.
Our results demonstrate that convolutions can perform at least on par with absolute position embeddings even in larger models.

\begin{table*}[t]
    \centering
    \small
    \renewcommand{\arraystretch}{1.2}
    \begin{tabular}{|>{\raggedright}p{4cm}|>{\raggedleft}p{1.1cm}|>{\raggedleft}p{0.8cm}>{\raggedleft}p{0.9cm}>{\raggedleft}p{0.9cm}>{\raggedleft}p{0.9cm}>{\raggedleft}p{0.8cm}>{\raggedleft}p{0.8cm}>{\raggedleft}p{0.75cm}>{\raggedleft}p{0.7cm}>{\raggedleft}p{0.7cm}|r|}
        \hline
        \textbf{Convolutions} & \textbf{Params} & CoLA & MNLI-m & MNLI-mm & MRPC & QNLI & QQP & RTE & SST & STS & \textbf{GLUE} \\
        \hline
        No convolution + abs position$^*$ & 13.43M & 30.8 & 76.1 & 75.9 & 80.4 & 78.5 & 74.4 & 62.2 & 85.1 & 76.8 & 71.1 \\
        \hline
        Composite (Equation \ref{eq:convq-convfixed}) & 13.43M & \textbf{40.4} & \textbf{78.2} & \textbf{77.4} & \textbf{85.0} & 83.3 & \textbf{77.7} & 64.7 & \textbf{87.8} & 82.1 & \textbf{75.2} \\
        %\hline
        Fixed depthwise & 13.47M & 36.9 & 77.6 & 76.1 & 80.6 & 81.9 & 76.4 & 64.5 & 87.5 & 79.7 & 73.5 \\
        Fixed depthwise + composite  & 13.48M & 38.0 & 77.4 & 76.3 & 82.8 & \textbf{83.7} & \textbf{77.7} & \textbf{65.3} & 87.3 & \textbf{82.3} & 74.5 \\
        \hline
    \end{tabular}
    \caption{GLUE test set performance for BERT-small models with added depthwise convolutions and composite attention. $^*$ denotes the default BERT-small model.}
    \label{non-lightweight-table}
\end{table*}

\section{Non-lightweight convolutions}
The previous section found that lightweight convolutions consistently improved pre-trained language model performance.
Next, we investigated whether the additional flexibility of non-lightweight convolutions could provide additional benefits.
Specifically, we considered convolutions that were fixed but non-lightweight.
In other words, convolution weights were fixed regardless of the input query, but weights were not tied across channels, equivalent to a standard depthwise convolution.
We only considered fixed depthwise convolutions because under existing frameworks, dynamic depthwise convolutions would introduce large numbers of parameters.

To implement depthwise convolutions, we added a convolution term identical to the fixed lightweight convolution in Equation \ref{eq:convfixed}, except that $\beta_{j-i}$ was learned separately for each feature channel:\footnote{For computational efficiency, we applied the softmax to the attention scores prior to adding the convolution term $\beta_{j-i, c}$, to avoid computing softmax scores separately for each individual channel. Softmax is not commonly applied in depthwise convolutions.}
\begin{equation}\label{eq:convolution}
\alpha_{ij, c} = \frac{(\textbf{x}_i W^Q) (\textbf{x}_j W^K)^T}{\sqrt{d_h}} + \beta_{j-i, c}
\end{equation}
This is equivalent to adding a depthwise convolution of the token values to the standard self-attention output.

\begin{figure}
    \centering
    % left, bottom, right, top
    \adjustbox{trim=0.5cm 0.5cm 0cm 0cm}{%
    \includegraphics[width=6cm]{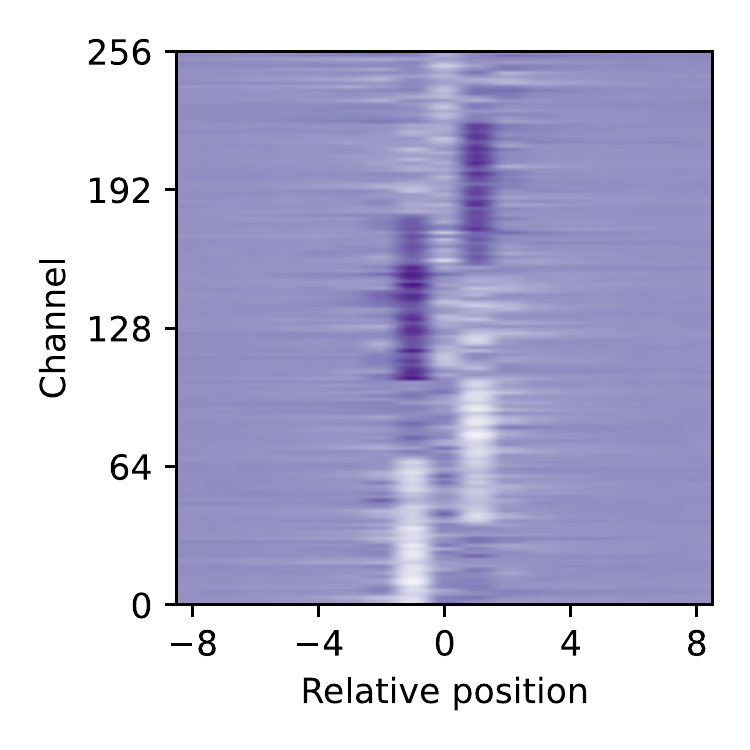}%
    }
    \caption{Learned convolution kernel weights $\beta_{j-i, c}$ (Equation \ref{eq:convolution}) for the depthwise convolution in the deepest attention layer.
    Channels correspond to the 256 features in each token representation.
    Channels are sorted such that kernels differentiating the previous and next token are grouped together.}
    \label{fig:convolution-kernel}
\end{figure}

\subsection{Non-lightweight convolution experiments}\label{non-lightweight-experiments}
We ran experiments using the same setup as the lightweight convolution experiments in Section \ref{section-lightweight-experiments}.
To compare the effects of dynamic lightweight convolutions (e.g. composite attention) and non-lightweight (depthwise) convolutions, we trained models using each possible combination of the two convolutions.
Results are presented in Table \ref{non-lightweight-table}.

\paragraph{Depthwise convolutions were less effective than lightweight convolutions.}
As with lightweight convolutions, the depthwise convolutions effectively replaced absolute position embeddings, outperforming the default model.
However, fixed depthwise convolutions performed worse than fixed lightweight convolutions on the majority of tasks.
This indicates that flexibility across channels is not critical to the success of convolutions in language models.

\begin{table*}[t]
    \rule{0pt}{10ex}  
    \centering
    \small
    \renewcommand{\arraystretch}{1.2}
    \begin{tabular}{|>{\raggedright}p{2cm}|>{\raggedright}p{2cm}|>{\raggedleft}p{1.5cm}|r|}
        \hline
        \multicolumn{4}{|c|}{ } \\[-0.8em]
        \multicolumn{4}{|c|}{\textbf{No composite attention}} \\[0.4em]
        \hline
        \textbf{Query/Key} & \textbf{Value} & \textbf{Params} & \textbf{GLUE} \\
        \hline
        Linear & Linear & 13.43M & $^*$71.1 \\
        Convolution & Linear & 13.53M & 71.9 \\
        Linear & Convolution & 13.47M & 73.4 \\
        Convolution & Convolution & 13.58M & 72.0 \\
        \hline
    \end{tabular}
    \hspace{0.5cm}
    \begin{tabular}{|>{\raggedright}p{2cm}|>{\raggedright}p{2cm}|>{\raggedleft}p{1.5cm}|r|}
        \hline
        \multicolumn{4}{|c|}{ } \\[-0.8em]
        \multicolumn{4}{|c|}{\textbf{+Composite attention}} \\[0.4em]
        \hline
        \textbf{Query/Key} & \textbf{Value} & \textbf{Params} & \textbf{GLUE} \\
        \hline
        Linear & Linear & 13.43M & \textbf{75.2} \\
        Convolution & Linear & 13.54M & 74.5 \\
        Linear & Convolution & 13.48M & 73.9 \\
        Convolution & Convolution & 13.59M & 74.0 \\
        \hline
    \end{tabular}
    \caption{BERT-small performance on the GLUE test set when replacing queries, keys, and values with depthwise-separable convolutions for half of the attention heads. $^*$ denotes the use of absolute position embeddings in the default BERT-small model.}
    \label{convolution-qkv-table}
\end{table*}

\paragraph{Composite attention already provided the necessary flexibility.}
Composite attention outperformed the fixed depthwise convolutions; even when composite attention was combined with depthwise convolutions, there was no overall improvement over composite attention alone.
This suggests that in the context of language, dynamic lightweight convolutions efficiently encode any local position information provided by depthwise convolutions.

\paragraph{Depthwise convolutions differentiated previous and next tokens.}
In previous sections, we found that lightweight convolution kernels often pay attention specifically to adjacent tokens.
As can be seen in Figure \ref{fig:convolution-kernel}, this result was even more pronounced in depthwise convolutions, with individual channels focusing on the previous or next token.
Interestingly, other channels specifically directed attention away from adjacent tokens.
This indicates that the relevant information about next and previous tokens can be compressed into a subset of the feature channels, freeing other channels to consider more distant or position-independent information.

\section{Convolutional queries, keys, and values}
Improvements over the non-convolutional baselines indicate that convolutions are beneficial to language model pre-training, serving as replacements for absolute position embeddings.
Our previous experiments applied different types of convolutions to self-attention values.
To take this result one step further, we replaced the linear query, key, and value projections themselves with convolutional layers.

Intuitively, applying convolutions before self-attention induces even more mixing of token representations.
If convolutions are built into every query, key, and value, then it becomes impossible for a token $i$ to pay attention to a single token $j$ without also incorporating information about tokens surrounding token $j$.

\subsection{Convolutional Q, K, V experiments}
As in Sections \ref{section-lightweight-experiments} and \ref{non-lightweight-experiments}, we ran experiments on BERT-small.
We replaced the query, key and value projections with depthwise-separable convolutions in half of the self-attention heads.\footnote{Depthwise-separable convolutions are a common way to save convolution parameters. A depthwise convolution is applied first, applying an independent convolution for each channel. Then, a pointwise convolution (i.e. a feedforward layer) mixes the channels to produce the final output.}
This aligns with previous work in which only half of the output dimensions for each token were generated using convolutions \citep{jiang-etal-2020-convbert}. Indeed, our initial explorations found that it was more effective to replace the linear projections in only half, not all, the attention heads.

Then, we considered whether convolutions from previous experiments provided additional benefits over convolutional queries, keys, and values.
To test this, we trained BERT-small models with composite attention (Equation \ref{eq:convq-convfixed}), adding convolutional queries, keys, and values.

\subsection{Convolutional Q, K, V results}
Results are presented in Table \ref{convolution-qkv-table}.
Similar to our previous convolution experiments, all convolutional replacements successfully outperformed the default model.
These results strongly support the conclusion that convolutions are a viable method of encoding positional information for language tasks.

However, all convolutional replacements for queries, keys, and values slightly decreased the performance of models using composite attention.
Convolutional values in particular were effective in models without composite attention, but they slightly decreased performance in models that already incorporated such lightweight convolutions.
We conclude that although convolutions can benefit models by adding local position information, there is a limit to how much local mixing should be done.
It is sufficient to apply convolutions to token values on top of self-attention; additional convolutional layers applied before the self-attention map enforce unnecessary mixing of token representations.

\section{Discussion}
Our results demonstrate that convolutions provide consistent benefits to pre-trained language models.
Our proposed composite attention mechanism combines previous relative position embedding methods, showing that convolutions can effectively compensate for the lack of local position information in Transformer models.

\subsection{Related work}
Our work unites and builds upon previous work using convolutions and relative positions in Transformers.
We adopted the relative embeddings from \citet{shaw-etal-2018-self} and \citet{huang-etal-2020-improve}, showing that these embeddings are equivalent to the dynamic lightweight convolutions in \citet{wu-etal-2019-pay}.
Combining these dynamic lightweight convolutions with fixed lightweight convolutions (equivalent to the relative position terms in \citealt{raffel-etal-2020-exploring}), we studied relative embeddings under the framework of convolution integrated with self-attention.
As far as we are aware, our work is the first to holistically compare relative positions, convolutions, and self-attention in language models.

Building upon dynamic lightweight convolutions, recent work has incorporated both depthwise-separable and dynamic lightweight convolutions in pre-trained language models. 
\citet{jiang-etal-2020-convbert} proposed ConvBERT, which adds a convolutional module alongside the standard self-attention mechanism in BERT.
ConvBERT's convolutional module consists of a depthwise-separable convolution combining with a query to generate a dynamic lightweight convolution.
Under our integrated framework, this is analogous to the model which uses depthwise-separable convolutions for queries and keys, using composite attention as a query-based dynamic lightweight convolution (see Table \ref{convolution-qkv-table}).
To make this comparison concrete, we trained a ConvBERT-small model using the same setup as our experiments.
Indeed, the analogous model under our framework outperformed ConvBERT-small (GLUE score 74.5 compared to 70.3).
Details for the ConvBERT comparison can be found in Appendix \ref{app:convbert}.

Finally, recent work has proved theoretical relationships between self-attention and convolution.
\citet{cordonnier-etal-2020-relationship} showed that given enough self-attention heads, self-attention weights can express any convolution; in fact, they showed that self-attention layers often learn such convolutional structures when trained on vision tasks.
However, this theoretical equivalence does not explain convolution-based improvements for Transformers in language tasks.
To clarify the relationship between self-attention and convolution in language, our work characterizes self-attention as a type of dynamic lightweight convolution.
By establishing a per-parameter equivalence between relative position embeddings and Wu's dynamic lightweight convolutions, we provide a concrete foundation where self-attention and convolution are used together in practice.

\section{Conclusion}
In this work, we formalized the relationship between self-attention and convolution.
We proposed composite attention, which combines self-attention with lightweight convolution, uniting previous approaches to relative positions.
Our formulation and empirical results demonstrate that convolutions can improve self-attention by providing local position information in sentences, capable of replacing absolute position embeddings entirely.

Our findings provide a solid foundation from which to study convolutions and self-attention in language tasks.
The spatially-oriented nature of convolutional neural networks translates directly into positional information in language.
As vision and language researchers strive towards common deep learning architectures, it is important to recognize how architectures for vision tasks can be adapted to linguistic domains.

\section*{Acknowledgments}
This work is funded by NSF IIS-1717431. Zhuowen Tu is also funded under the Qualcomm Faculty Award. Tyler Chang is partially supported by the UCSD HDSI graduate fellowship.

\bibliographystyle{acl_natbib}
\bibliography{anthology,acl2021}

\appendix
\section{Appendix}

\setlength\tabcolsep{3pt}
\begin{table}[h]
    \centering
    \small
    \renewcommand{\arraystretch}{1.2}
    \begin{tabular}{|>{\raggedright}p{3.5cm}|r|r|}
        \hline
        \textbf{Hyperparameter} & \textbf{Small} & \textbf{Base} \\
        \hline
        Layers & 12 & 12 \\
        Hidden size & 256 & 768 \\
        Intermediate hidden size & 1024 & 3072 \\
        Attention heads & 4 & 12 \\
        Attention head size & 64 & 64 \\
        Embedding size & 128 & 768\\
        Vocab size & 30004 & 30004 \\
        Max sequence length & 128 & 128 \\
        Mask proportion & 0.15 & 0.15 \\
        Learning rate decay & Linear & Linear \\
        Warmup steps & 10000 & 10000 \\
        Learning rate & 3e-4 & 1e-4 \\
        Adam $\epsilon$ & 1e-6 & 1e-6 \\
        Adam $\beta_1$ & 0.9 & 0.9 \\
        Adam $\beta_2$ & 0.999 & 0.999 \\
        Attention dropout & 0.1 & 0.1 \\
        Dropout & 0.1 & 0.1 \\
        Weight decay & 0.01 & 0.01 \\
        Batch size & 128 & 256 \\
        Train steps & 125K & 1M \\
        \hline
    \end{tabular}
    \caption{Pre-training hyperparameters.}
    \label{tab:pretrain-hyperparams}
\end{table}

\subsection{Pre-training and fine-tuning details}
\label{app:hyperparams}
BERT models (\citealt{devlin-etal-2019-bert}; \citealt{clark-etal-2020-electra}) were pre-trained on the BookCorpus \cite{zhu-etal-2015-aligning} and WikiText-103 datasets \cite{merity-etal-2017-pointer} using masked language modeling.
Pre-training examples were formatted as sentence pairs without the next sentence prediction objective.
In total, our dataset consisted of 31M unique sentence pairs.\footnote{Because BERT-small models were only trained for 125,000 steps with batch size 128, small models were trained on 16M sentence pairs.}
Sentences were tokenized by training an uncased SentencePiece tokenizer \cite{kudo-richardson-2018-sentencepiece}, and input and output token embeddings were tied during pre-training.
Models were evaluated on the GLUE benchmark \cite{wang-etal-2018-glue}.
Including ten fine-tuning runs for each GLUE task, each BERT-small model took about 24 hours to train on two Titan Xp GPUs.
Each BERT-base model took about 16 days to train on 8 GPUs.
Pre-training hyperparameters are listed in Table \ref{tab:pretrain-hyperparams}, and fine-tuning hyperparameters are listed in Table \ref{tab:finetune-hyperparams}.
Hyperparameters are based on those used in \citet{clark-etal-2020-electra} and \citet{devlin-etal-2019-bert}.

\begin{table}[t]
    \centering
    \small
    \renewcommand{\arraystretch}{1.2}
    \begin{tabular}{|>{\raggedright}p{3.4cm}|p{3.5cm}|}
        \hline
        \textbf{Hyperparameter} & \textbf{Value} \\
        \hline
        Learning rate decay & Linear \\
        Warmup steps & 10\% of total \\
        Learning rate & 1e-4 for QNLI or base-size \\
        & 3e-4 otherwise \\
        Adam $\epsilon$ & 1e-6 \\
        Adam $\beta_1$ & 0.9 \\
        Adam $\beta_2$ & 0.999 \\
        Attention dropout & 0.1 \\
        Dropout & 0.1 \\
        Weight decay & 0 \\
        Batch size & 128 for MNLI/QQP \\
        & 32 otherwise \\
        Train steps & 10 epochs for RTE/STS \\
        & 4 epochs for MNLI/QQP \\
        & 3 epochs otherwise \\
        \hline
    \end{tabular}
    \caption{Fine-tuning hyperparameters. We used intermediate task training for RTE, STS, and MRPC, initializing from a checkpoint fine-tuned on the MNLI task (\citealt{clark-etal-2020-electra}; \citealt{phang-etal-2018-sentence}).}
    \label{tab:finetune-hyperparams}
\end{table}

% START dev table.

\setlength\tabcolsep{3pt}
\begin{table*}[t]
    \centering
    \small
    \renewcommand{\arraystretch}{1.2}
    \begin{tabular}{|>{\raggedright}p{5cm}|>{\raggedleft}p{1.12cm}|>{\raggedleft}p{1.6cm}>{\raggedleft}p{1.6cm}>{\raggedleft}p{1.6cm}>{\raggedleft}p{1.6cm}r|}
        \hline
        \textbf{Convolution type, BERT-small} & \textbf{Params} & CoLA & MNLI-m & MNLI-mm & MRPC & \phantom{\_\_\_\_\_\_\_}QNLI \\
        \hline
        No convolution & 13.41M & $7.0 \pm 2.4$ & $73.0 \pm 0.1$ & $73.0 \pm 0.1$ & $80.9 \pm 0.4$ & $80.1 \pm 0.2$ \\
        No convolution + abs position$^*$ & 13.43M & $33.5 \pm 0.4$ & $75.8 \pm 0.1$ & $76.1 \pm 0.1$ & $83.3 \pm 0.4$ & $78.2 \pm 0.3$ \\
        \hline
        Fixed lightweight (\citealt{raffel-etal-2020-exploring}) & 13.42M & $38.3 \pm 0.8$ & $77.2 \pm 0.1$ & $77.2 \pm 0.1$ & $84.0 \pm 0.5$ & $82.1 \pm 0.1$ \\
        Dynamic lightweight (\citealt{shaw-etal-2018-self}) & 13.43M & $38.4 \pm 0.7$ & $\textbf{77.9} \pm 0.1$ & $77.6 \pm 0.1$ & $85.6 \pm 0.5$ & $82.8 \pm 0.1$ \\
        Composite (Equation \ref{eq:convq-convfixed}) & 13.43M & $\textbf{40.9} \pm 0.7$ & $\textbf{77.9} \pm 0.1$ & $78.0 \pm 0.1$ & $86.2 \pm 0.3$ & $83.0 \pm 0.1$ \\
        Composite + key-based dynamic & 13.44M & $40.0 \pm 0.6$ & $\textbf{77.9} \pm 0.1$ & $77.7 \pm 0.1$ & $\textbf{86.3} \pm 0.3$ & $83.3 \pm 0.1$ \\
        \hline
        Fixed depthwise & 13.47M & $38.0 \pm 0.6$ & $76.9 \pm 0.0$ & $76.8 \pm 0.1$ & $82.8 \pm 0.5$ & $81.9 \pm 0.1$ \\
        Composite + fixed depthwise & 13.48M & $40.4 \pm 0.7$ & $77.2 \pm 0.1$ & $77.4 \pm 0.1$ & $85.0 \pm 0.3$ & $83.3 \pm 0.1$ \\
        Convolutional QK & 13.53M & $33.4 \pm 0.4$ & $76.3 \pm 0.1$ & $76.4 \pm 0.1$ & $83.3 \pm 0.2$ & $81.3 \pm 0.2$ \\
        Convolutional value & 13.47M & $34.7 \pm 0.9$ & $76.2 \pm 0.0$ & $76.6 \pm 0.1$ & $83.4 \pm 0.4$ & $82.4 \pm 0.1$ \\
        Convolutional QKV & 13.58M & $31.9 \pm 0.7$ & $76.3 \pm 0.1$ & $76.3 \pm 0.1$ & $83.7 \pm 0.4$ & $80.4 \pm 0.2$ \\
        Composite + convolutional QK & 13.54M & $39.3 \pm 0.8$ & $77.4 \pm 0.1$ & $77.2 \pm 0.1$ & $85.4 \pm 0.3$ & $81.9 \pm 0.1$ \\
        Composite + convolutional value & 13.48M & $37.9 \pm 0.7$ & $77.8 \pm 0.1$ & $\textbf{78.1} \pm 0.1$ & $85.6 \pm 0.4$ & $\textbf{83.6} \pm 0.1$ \\
        Composite + convolutional QKV & 13.59M & $38.2 \pm 1.0$ & $77.4 \pm 0.1$ & $77.3 \pm 0.1$ & $85.3 \pm 0.4$ & $82.8 \pm 0.1$ \\
        ConvBERT & 13.09M & $33.3 \pm 1.5$ & $76.7 \pm 0.1$ & $76.8 \pm 0.1$ & $83.9 \pm 0.5$ & $77.1 \pm 0.8$ \\
        \hline
        \textbf{Convolution type, BERT-base} & & & & & & \\
        \hline
        No convolution + abs position$^*$ & 108.82M & $57.6 \pm 0.6$ & $\textbf{82.0} \pm 0.1$ & $\textbf{81.9} \pm 0.1$ & $\textbf{88.4} \pm 0.2$ & $84.7 \pm 0.3$ \\
        Fixed lightweight (\citealt{raffel-etal-2020-exploring}) & 108.73M & $\textbf{58.9} \pm 0.5$ & $81.9 \pm 0.1$ & $81.6 \pm 0.1$ & $87.7 \pm 0.3$ & $\textbf{86.2} \pm 0.1$ \\
        Dynamic lightweight (\citealt{shaw-etal-2018-self}) & 108.74M & $58.4 \pm 0.5$ & $81.8 \pm 0.1$ & $81.8 \pm 0.1$ & $86.7 \pm 0.4$ & $85.6 \pm 0.2$ \\
        Composite (Equation \ref{eq:convq-convfixed}) & 108.74M & $58.5 \pm 0.5$ & $81.9 \pm 0.1$ & $81.6 \pm 0.1$ & $86.0 \pm 1.2$ & $85.0 \pm 0.3$ \\
        \hline
    \end{tabular}
    
    \bigskip
    % TABLE PART 2:
    
    \begin{tabular}{|>{\raggedright}p{5cm}|>{\raggedleft}p{1.6cm}>{\raggedleft}p{1.6cm}>{\raggedleft}p{1.6cm}>{\raggedleft}p{1.6cm}r|}
        \hline
        \textbf{Convolution type, BERT-small} & QQP & RTE & SST & STS & \phantom{\_\_\_\_\_\_\_}\textbf{GLUE} \\
        \hline
        No convolution & $84.4 \pm 0.1$ & $61.0 \pm 0.5$ & $80.9 \pm 0.9$ & $83.7 \pm 0.1$ & $69.3 \pm 0.3$ \\
        No convolution + abs position$^*$ & $84.9 \pm 0.0$ & $64.4 \pm 0.5$ & $85.0 \pm 0.2$ & $82.4 \pm 0.1$ & $73.7 \pm 0.1$ \\
        \hline
        Fixed lightweight (\citealt{raffel-etal-2020-exploring}) & $86.2 \pm 0.0$ & $64.7 \pm 0.9$ & $86.9 \pm 0.2$ & $85.2 \pm 0.1$ & $75.7 \pm 0.2$ \\
        Dynamic lightweight (\citealt{shaw-etal-2018-self}) & $87.2 \pm 0.0$ & $65.1 \pm 0.9$ & $86.8 \pm 0.2$ & $85.6 \pm 0.1$ & $76.3 \pm 0.1$ \\
        Composite (Equation \ref{eq:convq-convfixed}) & $87.3 \pm 0.0$ & $66.1 \pm 0.7$ & $86.9 \pm 0.1$ & $85.9 \pm 0.1$ & $\textbf{76.9} \pm 0.1$ \\
        Composite + key-based dynamic & $87.4 \pm 0.0$ & $\textbf{66.3} \pm 0.4$ & $86.5 \pm 0.3$ & $86.1 \pm 0.2$ & $76.8 \pm 0.1$ \\
        \hline
        Fixed depthwise & $86.1 \pm 0.1$ & $64.2 \pm 0.7$ & $87.2 \pm 0.2$ & $84.4 \pm 0.1$ & $75.4 \pm 0.1$ \\
        Composite + fixed depthwise & $87.3 \pm 0.0$ & $63.5 \pm 0.8$ & $87.1 \pm 0.2$ & $86.1 \pm 0.1$ & $76.4 \pm 0.1$ \\
        Convolutional QK & $85.1 \pm 0.1$ & $63.0 \pm 1.0$ & $86.1 \pm 0.2$ & $84.5 \pm 0.1$ & $74.4 \pm 0.1$ \\
        Convolutional value & $86.6 \pm 0.0$ & $65.2 \pm 0.7$ & $87.2 \pm 0.3$ & $85.0 \pm 0.1$ & $75.2 \pm 0.1$ \\
        Convolutional QKV & $84.6 \pm 0.2$ & $66.1 \pm 0.9$ & $86.4 \pm 0.1$ & $84.4 \pm 0.1$ & $74.4 \pm 0.1$ \\
        Composite + convolutional QK & $86.7 \pm 0.0$ & $64.0 \pm 0.9$ & $\textbf{87.5} \pm 0.2$ & $85.7 \pm 0.1$ & $76.1 \pm 0.1$ \\
        Composite + convolutional value & $\textbf{87.5} \pm 0.0$ & $65.1 \pm 0.5$ & $\textbf{87.5} \pm 0.1$ & $\textbf{86.4} \pm 0.1$ & $76.6 \pm 0.1$ \\
        Composite + convolutional QKV & $87.0 \pm 0.0$ & $64.9 \pm 0.8$ & $86.9 \pm 0.1$ & $85.9 \pm 0.1$ & $76.2 \pm 0.2$ \\
        ConvBERT & $85.1 \pm 0.1$ & $64.6 \pm 0.5$ & $86.3 \pm 0.3$ & $84.0 \pm 0.2$ & $74.2 \pm 0.3$ \\
        \hline
        \textbf{Convolution type, BERT-base} & & & & & \\
        \hline
        No convolution + abs position$^*$ & $88.7 \pm 0.0$ & $69.9 \pm 0.5$ & $90.4 \pm 0.1$ & $\textbf{88.4} \pm 0.1$ & $81.0 \pm 0.2$ \\
        Fixed lightweight (\citealt{raffel-etal-2020-exploring}) & $\textbf{88.8} \pm 0.0$ & $70.9 \pm 0.7$ & $90.8 \pm 0.1$ & $88.1 \pm 0.1$ & $\textbf{81.3} \pm 0.2$ \\
        Dynamic lightweight (\citealt{shaw-etal-2018-self}) & $88.7 \pm 0.0$ & $70.6 \pm 0.6$ & $\textbf{91.1} \pm 0.1$ & $87.7 \pm 0.3$ & $81.1 \pm 0.2$ \\
        Composite (Equation \ref{eq:convq-convfixed}) & $88.7 \pm 0.0$ & $\textbf{71.0} \pm 0.7$ & $90.5 \pm 0.1$ & $\textbf{88.4} \pm 0.1$ & $81.2 \pm 0.2$ \\
        \hline
    \end{tabular}
    \caption{GLUE development set scores for each model described in the main paper, reporting averages and standard errors of the mean over ten fine-tuning runs for each task. $^*$ denotes the default BERT model.}
    \label{tab:dev-results}
\end{table*}

% END dev table.

\setlength\tabcolsep{3pt}
\begin{table*}[t]
    \centering
    \small
    \renewcommand{\arraystretch}{1.2}
    \begin{tabular}{|>{\raggedright}p{4cm}|>{\raggedleft}p{1.1cm}|>{\raggedleft}p{0.8cm}>{\raggedleft}p{0.9cm}>{\raggedleft}p{0.9cm}>{\raggedleft}p{0.9cm}>{\raggedleft}p{0.8cm}>{\raggedleft}p{0.8cm}>{\raggedleft}p{0.75cm}>{\raggedleft}p{0.7cm}>{\raggedleft}p{0.7cm}|r|}
        \hline
        \textbf{Model, BERT-small} & \textbf{Params} & CoLA & MNLI-m & MNLI-mm & MRPC & QNLI & QQP & RTE & SST & STS & \textbf{GLUE} \\
        \hline
        ConvBERT & 13.1M & 25.5 & 75.4 & 73.9 & 79.7 & 76.0 & 74.7 & 64.3 & 85.6 & 77.9 & 70.3 \\
        \hline
        Integrated convolutions and self-attention (ours) & 13.5M & \textbf{37.9} & \textbf{77.5} & \textbf{76.6} & \textbf{83.7} & \textbf{83.1} & \textbf{76.6} & \textbf{65.3} & \textbf{88.7} & \textbf{81.1} & \textbf{74.5} \\
        \hline
    \end{tabular}
    \caption{Comparison between ConvBERT-small and the analogous model under our framework, reporting GLUE test set results.}
    \label{tab:convbert-table}
\end{table*}

\subsection{GLUE development results}
\label{app:dev-results}
Results for each model on the GLUE development set are reported in Table \ref{tab:dev-results}.
We report averages over ten fine-tuning runs for each task, including standard errors of the mean.
Each overall GLUE score was computed as the average of individual task scores; we computed GLUE score averages and standard errors over ten GLUE scores, corresponding to the ten fine-tuning runs.
We note that development scores were generally higher than test scores due to differences between the test and training distributions \cite{wang-etal-2018-glue}.

\subsection{Detailed ConvBERT comparison}
\label{app:convbert}
ConvBERT adds a convolutional module alongside the standard self-attention mechanism in BERT \cite{jiang-etal-2020-convbert}.
ConvBERT uses half the number of standard self-attention heads, using convolutional modules for the other half.
In each convolutional module, a depthwise-separable convolution is multiplied pointwise with the query in the corresponding self-attention head. This convolutional query is fed into a linear layer to generate a dynamic lightweight convolution.

Under our framework, the analogous model replaces half of the queries and keys with depthwise-separable convolutions and uses composite attention (a query-based dynamic lightweight convolution; see Table \ref{convolution-qkv-table} in the full paper).
In both models (ConvBERT and our own), half of the attention heads use a convolutional query.
Additionally, in both models, the convolutional query is used to generate a dynamic lightweight convolution.

However, in our model, the dynamic lightweight convolution (in this case, composite attention) is used for all attention heads, not just the convolutional heads.
Furthermore, our convolutional heads still use a self-attention mechanism along with the dynamic lightweight convolutions, by generating convolutional keys.
In this way, our model adds convolutions to ConvBERT's self-attention heads, and adds self-attention to ConvBERT's convolutional heads.

Then, we investigated whether the separate self-attention and convolutional modules in ConvBERT provide any benefit over our integrated convolution and self-attention.
We trained a ConvBERT-small model using the same pre-training setup as our BERT-small experiments, comparing performance to the analogous model under our framework.
Results are shown in Table \ref{tab:convbert-table}.
Indeed, integrated convolutions and self-attention outperformed ConvBERT-small, using only 3\% more parameters.

\end{document}